\documentclass[conference]{IEEEtran}
\IEEEoverridecommandlockouts

\usepackage{cite}
\usepackage{amsmath,amssymb,amsfonts}
\usepackage{algorithmic}
\usepackage{graphicx}
\usepackage{textcomp}
\usepackage{xcolor}
\usepackage{subcaption} 
\usepackage{booktabs}   
\usepackage{tikz}       
\usepackage{pgfplots}   
\pgfplotsset{compat=1.18} 
\usepackage{orcidlink}

\def\BibTeX{{\rm B\kern-.05em{\sc i\kern-.025em b}\kern-.08em
    T\kern-.1667em\lower.7ex\hbox{E}\kern-.125emX}}

\begin{document}


\title{Impact of Data-Oriented and Object-Oriented Design on Performance and Cache Utilization with Artificial Intelligence Algorithms in Multi-Threaded CPUs
\thanks{The authors would like to thank FAPERGS (24/2551-0001396-2, 23/2551-0000773-8), CNPq (305805/2021-5) and FAPERGS/CNPq (23/2551-0000126-8).}
}

\author{
    \IEEEauthorblockN{
        Gabriel M. Arantes\orcidlink{0009-0002-7296-2607}\IEEEauthorrefmark{1},
        Giancarlo Lucca\orcidlink{0000-0002-3776-0260}\IEEEauthorrefmark{2},
        Eduardo N. Borges\orcidlink{0000-0003-1595-7676}\IEEEauthorrefmark{1},
        Richard F. Pinto\orcidlink{0009-0007-0176-3383}\IEEEauthorrefmark{1},
        Bruno L. Dalmazo\orcidlink{0000-0002-6996-7602}\IEEEauthorrefmark{1} \\ and
        Rafael A. Berri\orcidlink{0000-0002-3812-4186}\IEEEauthorrefmark{1}
    }
    \IEEEauthorblockA{
        \IEEEauthorrefmark{1}\textit{Federal University of Rio Grande (FURG), Rio Grande, Brazil} \\
        \IEEEauthorrefmark{2}\textit{Catholic University of Pelotas (UCPel), Pelotas, Brazil}
    }
}

\maketitle

\begin{abstract}
The growing performance gap between multi-core CPUs and main memory necessitates hardware-aware software design paradigms. This study provides a comprehensive performance analysis of Data-Oriented Design (DOD) versus the traditional Object-Oriented Design (OOD), focusing on cache utilization and efficiency in multi-threaded environments. We developed and compared four distinct versions of the A* search algorithm: single-threaded OOD (ST-OOD), single-threaded DOD (ST-DOD), multi-threaded OOD (MT-OOD), and multi-threaded DOD (MT-DOD). The evaluation was based on metrics including execution time, memory usage, and CPU cache misses. In multi-threaded tests, the DOD implementation demonstrated considerable performance gains, with faster execution times and a lower number of raw system calls and cache misses. While OOD occasionally showed marginal advantages in memory usage or percentage-based cache miss rates, DOD's efficiency in data-intensive operations was more evident. Furthermore, our findings reveal that for a fine-grained task like the A* algorithm, the overhead associated with thread management led to single-threaded versions significantly outperforming their multi-threaded counterparts in both paradigms. We conclude that even when performance differences appear subtle in simple algorithms, the consistent advantages of DOD in critical metrics highlight its foundational architectural superiority, suggesting it is a more effective approach for maximizing hardware efficiency in complex, large-scale AI and parallel computing tasks.
\end{abstract}



\section{Introduction}

In the modern world, computers are extremely important and are used in nearly every aspect of daily life. They are continuously advancing, with constant improvements in their performance and capabilities, making their efficient use increasingly necessary and important \cite{NRC2011}.

In particular, the CPU (\textit{Central Processing Unit}) is evolving at a rapid pace, with annual improvements in processing speed and memory storage, known in the CPU context as cache. The cache has different levels of speed and size, whereas other components have not evolved at the same rate \cite{nyberg2021investigating}. Storage speed outside the CPU is significantly slower, and in this paper, we will focus on RAM, which is approximately 100 times slower than the CPU \cite{carvalho2002gap}.

A solution to mitigate this performance gap may lie in the more effective use of the cache through the adoption of Data-Oriented Design/Data-Oriented Programming (DOD/DOP) patterns, as opposed to the current industry standard of Object-Oriented Design/Object-Oriented Programming (OOD/OOP).

OOP is not efficient in terms of cache usage, as it has a greater impact on performance and scalability, whereas DOD does not exhibit these issues \cite{nikolov2018oop}. OOP consists of using object classes for data manipulation and well-encapsulated functions with multiple abstraction layers \cite{stroustrup1988object}. In contrast, DOD makes better use of data by separating it from the code, thus improving data access and allowing for greater scalability and easier code maintenance \cite{joshi2007data}.

The remainder of the paper is organized as follows. Section II presents a literature review. The methodology is presented in Section III, and details of its results in Section IV. Finally, in Section V, some final remarks are made, and directions for future research are indicated.

\section{Literature Review} \label{sec:firstpage}

To develop this paper, a study on the subject was conducted by reviewing scientific literature, and this section presents the results of that study.

\subsection{Articles and Books}

Computer architecture is a vast and constantly evolving field, with numerous significant contributions over the decades. In this context, the book "Computer Architecture" by Hennessy and Patterson \cite{patterson1988} highlights the importance of memory in computing, particularly the role of cache in enhancing overall system performance. They argue that systems with more efficient caches tend to perform better overall, even though they do not provide a study specifically focused on cache, supporting their assertion through a broad analysis of computer architectures.

The relevance of cache is corroborated by various other studies and publications. For example, Culler et al. \cite{culler1998}, in "Parallel Computer Architecture: A Hardware/Software Approach," emphasize that the performance of parallel systems is crucial to avoiding bottlenecks and demonstrate how effective cache utilization can significantly improve system performance. Similarly, Drepper \cite{drepper2007}, in "What Every Programmer Should Know About Memory," discusses how efficient cache management can reduce memory access time, thereby improving program performance.

Leslie Lamport, in his seminal paper "Time, Clocks, and the Ordering of Events in a Distributed System" \cite{lamport1978}, introduces concepts of continuity and synchronization in distributed systems, emphasizing the importance of cache in reducing latency and enhancing performance. This perspective is complemented by Gene Amdahl \cite{amdahl1967}, who mentions the significance of cache in minimizing memory access latency in his work on Amdahl's Law, although his primary focus is on parallelism.

The impact of cache on performance is also addressed by Patterson et al. \cite{patterson1988} in "A Case for Redundant Arrays of Inexpensive Disks (RAID)," where they discuss how efficient caches can improve the performance of data storage systems. Dean and Ghemawat \cite{dean2004}, in "MapReduce: Simplified Data Processing on Large Clusters," highlight the importance of intelligent cache usage to enhance the performance of processing nodes in clusters, which is essential for the parallel processing of large datasets.

In "High-Performance Concurrency Control Mechanisms for Main-Memory Databases," Larson et al. \cite{larson2011} analyze the use of multi-threading to increase cache efficiency in highly concurrent database environments, demonstrating that an efficient cache is crucial for reducing latency and increasing throughput. Similarly, Bienia et al. \cite{bienia2008}, in "The PARSEC Benchmark Suite: Characterization and Architectural Implications," present benchmarks that show how the structure and effectiveness of the cache are fundamental to achieving good performance in parallel applications dealing with large volumes of data.

While Object-Oriented Programming (OOP) has been widely adopted in the software industry, its limitations have led to the exploration of alternative approaches, such as Data-Oriented Design (DOD). DOD focuses on data organization and efficiency, in contrast to OOP, which emphasizes structuring code into objects and classes.

The primary advantage of DOD is efficiency. By organizing data in a way that aligns with hardware memory architecture, DOD can minimize cache misses and maximize cache efficiency. This is particularly important in performance-critical systems, such as gaming, simulations, and scientific applications. DOD emphasizes memory contiguity, enabling fast and efficient data access, reducing latency, and increasing data throughput.

One of the most vocal critics of OOP is Joe Armstrong, the creator of the Erlang programming language. In his article "Why OO Sucks" \cite{armstrong2003}, Armstrong argues that the strong interdependence between objects and the complexity of class hierarchies can lead to rigid systems that are difficult to modify. He asserts that DOD, by focusing on data structure and flow, offers greater flexibility and simplicity. This approach can result in systems that are easier to maintain and scale, as the programmer's logic becomes clearer and more direct.

Paul Graham, in "Hackers \& Painters" \cite{graham2004}, also criticizes OOP for the performance overhead it can cause. He argues that abstraction and encapsulation, fundamental to OOP, can perform worse compared to more direct approaches such as procedural programming or DOD. The simplicity of DOD can lead to more efficient and easier-to-understand code, avoiding the excessive complexity that often accompanies OOP.

Robert C. Martin, in "Clean Code" \cite{martin2009}, acknowledges that OOP can create unnecessary complexity if not used cautiously. While he does not completely discard OOP, Martin suggests that maintaining simplicity and avoiding complex class hierarchies are essential practices. DOD, by its nature, tends to avoid these problems by focusing on data structures in a more linear and direct manner.

Therefore, while OOP offers advantages such as encapsulation and code reuse, its disadvantages, particularly in terms of performance and complexity, have led to a growing adoption of DOD in contexts where efficiency is paramount. DOD, by aligning data organization with hardware architecture, provides a simpler and more efficient approach, facilitating system maintenance and scalability. Moreover, cache efficiency is widely recognized as crucial for the performance of computing systems, being a focal area in various studies and publications over the years. The combination of DOD techniques with efficient cache management can result in highly performant systems that are adaptable to modern data processing needs.

\subsection{Related Work}

The article by Pellacini et al. \cite{pellacini2019yocto}, "Yocto/GL: A Data-Oriented Library For Physically-Based Graphics," emphasizes the use of Data-Oriented Design (DOD) as a way to achieve simplicity, composability, and scalability in computer graphics. The authors contrast this approach with Object-Oriented Design (OOD), frequently used in other libraries, which they argue can lead to complexity and performance issues, especially when handling large datasets. Specifically, Yocto/GL uses free functions that operate on basic data types, value semantics for assignments, and index references instead of pointers. These design choices aim to improve code readability, facilitate the composition of different functionalities, and promote a more memory-efficient data layout.

In "Performance of Data-Oriented Design with Unity DOTS," Venâncio \cite{venancio2023desempenho} compares Data-Oriented Design (DOD) with Object-Oriented Design (OOD) in game development using the Unity engine and the DOTS package. The author argues that DOD, by focusing on processing and data, facilitates the implementation of multi-threading, resulting in better CPU utilization and consequently improved performance. The study includes a tank battle simulator, "Metal Rain," to test both approaches. The results show that DOD, with DOTS, achieves higher frames per second (FPS) and better CPU core utilization, especially in scenarios with many elements. It is important to note that the author observes that in cases of GPU bottlenecks, the benefits of DOD in terms of FPS may be less apparent. Although the study does not delve deeply into cache optimization, the author mentions that DOD generally organizes data in arrays, which favors cache memory usage.

In "Exploiting Cache Locality to Speedup Register Clustering," Fontana et al. \cite{fontana2017exploiting} argue that data organization directly influences algorithm performance, especially in tasks such as register clustering, which involve large volumes of data. The authors compare Data-Oriented Design (DOD) with traditional Object-Oriented Design (OOD), demonstrating that DOD, by organizing data into separate arrays per property, reduces cache space wastage and improves cache spatial locality. This optimization translates into fewer cache misses and consequently faster execution times in both sequential and parallel (multi-threaded) implementations. The article demonstrates, through experiments with the K-means algorithm, that the data organization proposed by DOD accelerates processing, particularly in high-memory-use scenarios, making it a promising approach for Electronic Design Automation (EDA) tools, which are software tools used to design and produce electronic systems, from computer chips to printed circuit boards.

In "Investigating the Effect of Implementing Data-Oriented Design Principles on Performance and Cache Utilization," Nyberg et al. \cite{nyberg2021investigating} argue that Data-Oriented Design (DOD) principles can significantly enhance performance in game development environments, primarily through optimized CPU cache utilization. The author explores the removal of runtime polymorphism, iteration over contiguous data, and reduction of data size as core DOD principles that directly influence cache performance. Through experiments with a test program simulating movement and collision detection, Nyberg demonstrates that these DOD optimizations result in faster CPU times and lower cache miss rates compared to traditional object-oriented designs. The study highlights the importance of data locality and efficient cache usage for high-performance games, especially in data-intensive scenarios. However, the author acknowledges that the scope of the research is limited to single-threaded tests and suggests further investigation in multi-threaded environments as a topic for future research.

Wingqvist et al. \cite{wingqvist2022}, in "Evaluating the Performance of Object-Oriented and Data-Oriented Design with Multi-Threading in Game Development," developed a simple game using two programming design patterns, Object-Oriented Design (OOD) and Data-Oriented Design (DOD), to compare their performance in terms of runtime and cache utilization, both in single-threaded and multi-threaded scenarios. They ran the game on different systems with various numbers of cores and cache sizes and found that DOD consistently outperformed OOD across all configurations, exhibiting significantly faster execution times (achieving a speedup of up to 13.25 times) and lower cache miss rates (with an improvement of up to 5.57 times in multi-threaded scenarios). The authors concluded that DOD is particularly beneficial for game development, especially on modern multi-core systems, as it utilizes CPU resources more efficiently. However, they acknowledged that DOD is more complex to implement than OOD and suggested future research with large-scale games to further validate their findings.

The research approach in this article will stand out by fully integrating cache and multi-threading principles into a single study focused on parallel computing systems. While Nyberg et al. and Wingqvist et al. address these two topics separately, this research aims to investigate how the intersection between effective caching and multi-threading can be optimized to enhance parallel system performance under different load conditions and configurations. This will be tested and examined using both OOD and DOD programming paradigms to assess their efficiency in this regard. The impact of these technologies will be analyzed, proposing new models and approaches with the potential to yield significant computational performance improvements in parallel and distributed environments.

\section{Methodology}

To achieve the objectives of this paper, a simple artificial intelligence algorithm was developed using two different programming paradigms. Additionally, the algorithm was parallelized, resulting in four versions: single-threaded OOD (ST-OOD), single-threaded DOD (ST-DOD), multi-threaded OOD (MT-OOD), and multi-threaded DOD (MT-DOD).

Various parameters were considered, such as the number of threads used, the size, and the complexity of the data manipulated, to ensure a comprehensive comparative analysis between the approaches.

The test program was developed in Python by the author, selecting this language for its simplicity and rapid development, with an accessible syntax and powerful data manipulation tools, in addition to various libraries such as "pyamaze," which was widely used in this algorithm for generating mazes for the test program.

The experiment consisted of implementing the A* (\textit{A star}) algorithm, a simple search algorithm used to find the shortest path between two points in a graph or map. As shown in Figure 1, the algorithm started from the bottom-right corner and found the optimal path to the top-left corner. The algorithm's operation can be summarized as follows:

\begin{figure*}[t] 
\centering
\includegraphics[width=0.8\textwidth]{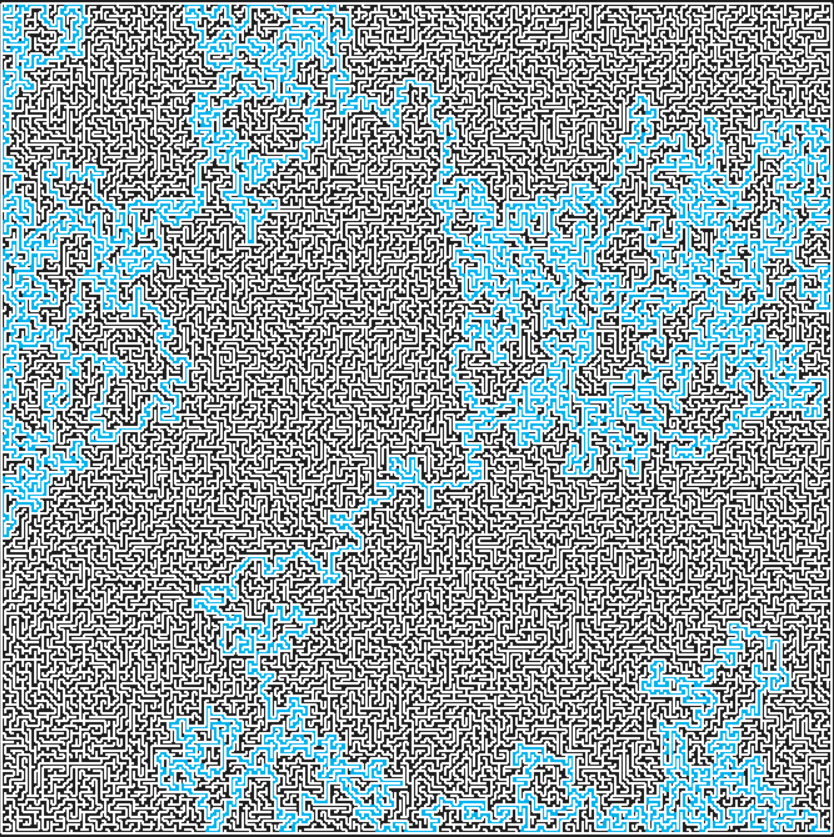}
\caption{Example of the A* algorithm in execution}
\label{fig:barplot}
\end{figure*}

\begin{itemize}
    \item Each node was associated with essential information for the execution of A*: accumulated cost (g), estimated cost to the goal (h), and total cost (f = g + h).
    \item To find the shortest path from the start node to the goal, the algorithm maintains an open list and a closed list:
    \begin{enumerate}
        \item The node with the lowest total cost (f) is selected from the open list.
        \item The selected node's neighbors are evaluated: for each neighbor, its accumulated cost (g) and estimated cost (h) are calculated, and the total cost (f) is updated.
        \item If a neighbor is not yet in the open list, it is added; otherwise, the algorithm checks whether the new path is more efficient (lower g).
        \item The processed node is moved to the closed list.
    \end{enumerate}
    \item The algorithm terminates when the goal node is reached or when there are no more nodes in the open list.
\end{itemize}

The effectiveness of the A* algorithm is attributed to its heuristic function, denoted as 'h'. This function provides an informed estimate of the cost from any given node to the goal. Unlike uninformed search algorithms such as Breadth-First Search (BFS) or Depth-First Search (DFS), which explore paths without guidance towards the target, A* uses the heuristic to prioritize nodes, significantly reducing the search space. In this implementation, the "Manhattan Distance" was employed as the heuristic function. This calculates the sum of the absolute differences of the x and y coordinates between the current cell and the destination cell. Mathematically, for a current cell $(x_c, y_c)$ and a destination cell $(x_d, y_d)$, the Manhattan Distance $h$ is given by:

$$h = |x_c - x_d| + |y_c - y_d|$$

This particular heuristic is admissible, meaning it never overestimates the true cost to reach the goal, which is a crucial property for A* to guarantee finding the optimal path. The use of an admissible heuristic, like Manhattan Distance in a grid-based maze, ensures that the algorithm explores promising paths first, making it considerably more efficient than its uninformed counterparts by guiding the search directly towards the objective.

\subsection*{Structure of the A* Algorithm in OOD}

For the experiments using OOD, the structure followed a class-based design pattern, utilizing an AoS (Array of Structs):

\begin{verbatim}
Class Celula
    Float linha
    Float coluna
    Float g_score
    Float f_score

Class AStarDOD
    Array mapa
    Tuple inicio
    Tuple destino
    Dictionary celulas
    Dictionary caminho
\end{verbatim}

\subsection*{Structure of the A* Algorithm in DOD}

For the experiment using DOD, the design pattern used was different, storing data not as classes but as an SoA (Struct of Arrays):

\begin{verbatim}
Struct Celula
    Float linha
    Float coluna
    Float g_score
    Float f_score

Function main()
    linhas: int
    colunas: int
    destino: Tuple[int, int]
    
\end{verbatim}

In this way, object information was stored sequentially in memory, which is expected to optimize the tests.

For the implementation of the codes, two libraries were used: "pyamaze" for maze generation and "PriorityQueue" to use a priority queue in the algorithm. Additionally, in the two multi-threaded codes, the "concurrent.futures" library was used to assist with multi-threading.

\begin{center}
\begin{table}[h]
    \centering
    \begin{tabular}{c c c}
        \toprule
         & Name          & Dell Inspiron 15-7579 \\
         & CPU           & Intel Core i5-7200U \\ 
         & CPU Freq GHz  & 2.5 (Turbo Boost up to 3.1) \\
         & Cores         & 2 \\
         & Threads       & 4 \\
         & L1 Cache Kb   & 128 \\
         & L2 Cache Kb   & 512 \\
         & L3 Cache Mb   & 3 \\
         & RAM           & DDR4 16 GB \\
         & RAM Freq MHz  & 2133 \\
         & OS            & Ubuntu 24.04.1 LTS \\
         \bottomrule \\
    \end{tabular}
    \caption{System used for empirical tests.}
    \label{tab:my_label}
\end{table}
\end{center}

Additionally, parameter selection was analyzed to maximize the author's machine performance (Table \ref{tab:my_label}), including the number of threads used, and the size and complexity of the manipulated data.

Thus, the codes used a pre-built maze downloaded with a width and height of 200, generating 40,000 cells, where the starting point is (40000,40000) and the destination is (1,1). The same maze was used across all tests to avoid bias in the experiment. Furthermore, the maze used is a perfect maze, meaning there are no dead ends.

\section{Results}

For evaluation, the collected data were analyzed in three aspects: execution time, memory usage, cache misses, and, in multi-threaded codes, the number of threads used. These metrics are used to compare the behavior of the test program with OOP and DOD design patterns in a parallel execution context with four threads on the operating system.

For this research, all experiments were conducted on Ubuntu 24.04.1 LTS, utilizing four threads for parallel execution to assess performance metrics.

\subsection{Sequential Execution}

Sequential execution refers to the single-thread execution of the test program, meaning a single thread worked sequentially to solve the algorithm.

\subsubsection{Execution Time}

Figure \ref{fig:sequential_performance_et_mu}(\subref{fig:sequential_performance_et}) compares the sequential performance of the test program using DOD and OOD. The y-axis shows the execution time in seconds, while the values above the bars indicate the performance improvement (speedup) of ST-DOD W compared to the ST-OOD W implementation.

As shown in the graph, DOD exhibits slightly better performance than OOD in the single-thread implementation. The execution time of the OOD code was 2.981 seconds, while for DOD, it was 2.791 seconds. The observed improvement was approximately 1.068 times when executing the test program on Ubuntu 24.04.1 LTS with four threads using 40,000 cells.

\subsubsection{Memory Usage}

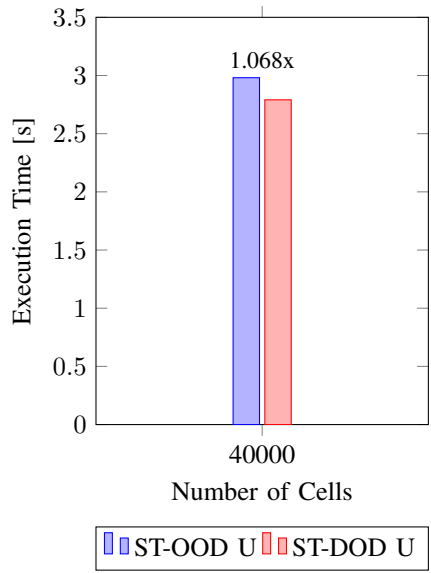
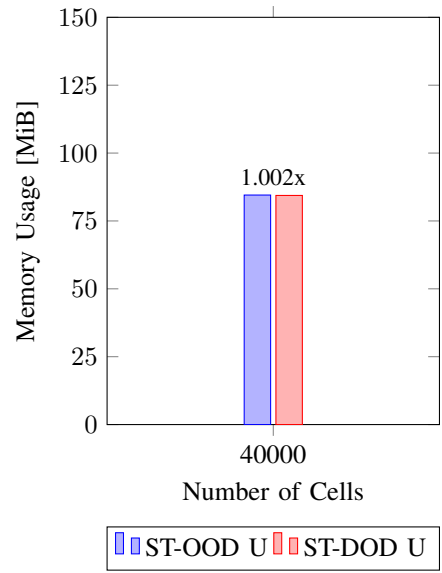
\begin{figure*}[h]
    \centering
    \begin{minipage}[t]{0.48\textwidth}
        \centering
        \begin{tikzpicture}
            \begin{axis}[
                ybar,
                bar width=10pt,
                symbolic x coords={40000},
                xtick=data,
                ymin=0, ymax=3.5,
                ylabel={Execution Time [s]},
                xlabel={Number of Cells},
                width=6cm,
                height=7cm,
                ytick={0, 0.5, 1, 1.5, 2, 2.5, 3, 3.5},
                enlarge x limits={abs=0.75cm},
                legend style={at={(0.5,-0.25)}, anchor=north, legend columns=-1},
            ]
            \addplot coordinates {(40000, 2.981)};
            \addplot coordinates {(40000, 2.791)};
            \legend{ST-OOD U, ST-DOD U}

            \node at (axis cs:40000, 2.981) [above, font=\small] {1.068x}; 
            \end{axis}
        \end{tikzpicture}
        \subcaption{Execution time graph generated on Ubuntu system.}
        \label{fig:sequential_performance_et}
    \end{minipage}
    \hfill
    \begin{minipage}[t]{0.48\textwidth}
        \centering
        \begin{tikzpicture}
            \begin{axis}[
                ybar,
                bar width=10pt,
                symbolic x coords={40000},
                xtick=data,
                ymin=0, ymax=150,
                ylabel={Memory Usage [MiB]},
                xlabel={Number of Cells},
                width=6cm,
                height=7cm,
                ytick={0, 25, 50, 75, 100, 125, 150},
                enlarge x limits={abs=0.75cm},
                legend style={at={(0.5,-0.25)}, anchor=north, legend columns=-1},
            ]
            \addplot coordinates {(40000, 84.539)};
            \addplot coordinates {(40000, 84.394)};
            \legend{ST-OOD U, ST-DOD U}

            \node at (axis cs:40000, 84.539) [above, font=\small] {1.002x}; 
            \end{axis}
        \end{tikzpicture}
        \subcaption{Memory usage graph for the test conducted on Ubuntu system.}
        \label{fig:sequential_performance_mu}
    \end{minipage}
    \caption{Single-thread: execution time and memory usage.}
    \label{fig:sequential_performance_et_mu}
\end{figure*}

Figure \ref{fig:sequential_performance_et_mu}(\subref{fig:sequential_performance_mu}) compares the memory usage of the test program using DOD and OOD. The y-axis represents the amount of memory used by the algorithm in mebibytes ($ 2^{20} $ bytes), while the values above the bars indicate the additional memory used by the OOD implementation compared to DOD.

As shown in the graph, DOD again exhibits slightly better performance than OOD in the single-thread implementation. The memory usage of the OOD code was 84.539 MiB, while for DOD, it was 84.394 MiB. The observed improvement was approximately 1.002 times when executing the test program on Ubuntu 24.04.1 LTS with four threads using 40,000 cells.

\subsubsection{Cache Utilization}

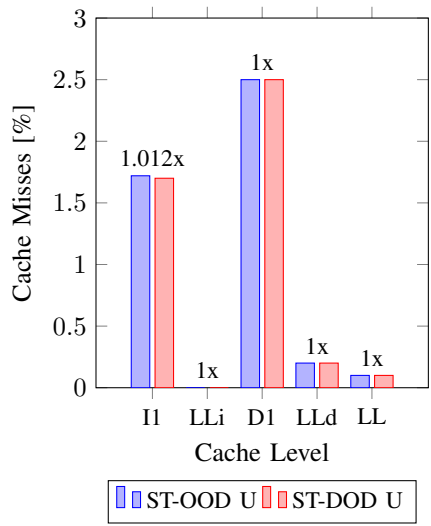
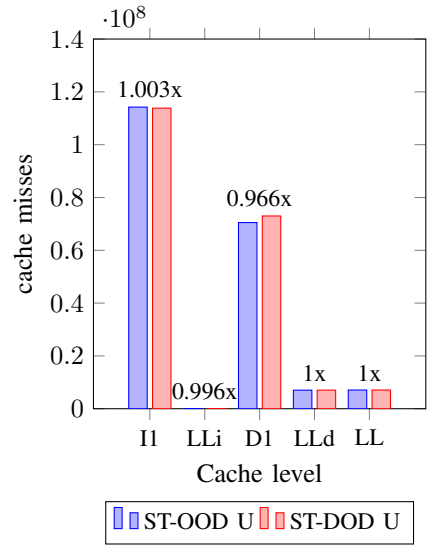
\begin{figure*}[h]
    \centering
    \begin{minipage}[t]{0.48\textwidth}
        \centering
        \begin{tikzpicture}
            \begin{axis}[
                ybar,
                bar width=7pt, 
                symbolic x coords={I1, LLi, D1, LLd, LL},
                xtick=data,
                ymin=0, ymax=3,
                ylabel={Cache Misses [\%]},
                xlabel={Cache Level},
                width=6cm, 
                height=6.5cm,
                ytick={0, 0.5, 1, 1.5, 2, 2.5, 3},
                enlarge x limits={abs=0.75cm},
                legend style={at={(0.5,-0.25)}, anchor=north, legend columns=-1, font=\small},
                xticklabel style={font=\small, text width=0.8cm, align=center},
            ]
            \addplot coordinates {(I1, 1.72) (LLi, 0) (D1, 2.5) (LLd, 0.2) (LL, 0.1)};
            
            \addplot coordinates {(I1, 1.70) (LLi, 0) (D1, 2.5) (LLd, 0.2) (LL, 0.1)};
            
            \legend{ST-OOD U, ST-DOD U}

            \node at (axis cs:I1, 1.72) [above] {\small 1.012x}; 
            \node at (axis cs:LLi, 0.0) [above] {\small 1x}; 
            \node at (axis cs:D1, 2.5) [above] {\small 1x};
            \node at (axis cs:LLd, 0.2) [above] {\small 1x};
            \node at (axis cs:LL, 0.1) [above] {\small 1x};
            \end{axis}
        \end{tikzpicture}
        \subcaption{Cache misses [\%] graph generated on the Ubuntu system.}
        \label{fig:sequential_cache_cu}
    \end{minipage}
    \hfill
    \begin{minipage}[t]{0.48\textwidth}
        \centering
        \begin{tikzpicture}
            \begin{axis}[
                ybar,
                bar width=7pt, 
                symbolic x coords={I1, LLi, D1, LLd, LL},
                xtick=data,
                ymin=0, ymax=140000000,
                ylabel={cache misses},
                xlabel={Cache level},
                width=6cm, 
                height=6.5cm,
                ytick={0, 20000000, 40000000, 60000000, 80000000, 100000000, 120000000, 140000000},
                enlarge x limits={abs=0.75cm},
                legend style={at={(0.5,-0.25)}, anchor=north, legend columns=-1, font=\small},
                xticklabel style={font=\small, text width=0.8cm, align=center},
            ]
            \addplot coordinates {(I1, 114226091) (LLi, 57208) (D1, 70495485) (LLd, 7032268) (LL, 7089476)};
            
            \addplot coordinates {(I1, 113838916) (LLi, 57414) (D1, 73000810) (LLd, 7032341) (LL, 7089755)};
            
            \legend{ST-OOD U, ST-DOD U}

            \node at (axis cs:I1, 114226091) [above] {\small 1.003x};
            \node at (axis cs:LLi, 57414) [above] {\small 0.996x};
            \node at (axis cs:D1, 73000810) [above] {\small 0.966x};
            \node at (axis cs:LLd, 7032341) [above] {\small 1x};
            \node at (axis cs:LL, 7089755) [above] {\small 1x};
            \end{axis}
        \end{tikzpicture}
        \subcaption{Graph of raw cache misses generated in the Ubuntu system.}
        \label{fig:sequential_cache_rm}
    \end{minipage}
    \caption{Single-thread: cache misses and raw cache misses.}
    \label{fig:sequential_cache_cu_rm}
\end{figure*}

Figure \ref{fig:sequential_cache_cu_rm}(\subref{fig:sequential_cache_cu}) presents the performance results related to cache misses of the sequential implementations. It highlights the difference in cache miss rates between ST-DOD U (Ubuntu) and ST-OOD U.

The y-axis shows the percentage of cache misses, while the values above the bars indicate the additional cache misses observed in the OOD implementation compared to DOD.

The x-axis details the cache levels analyzed in terms of cache misses. These levels include: I1, LLi, D1, LLd, and LL. I1 represents the first-level instruction cache, LLi is the last level of the instruction cache, D1 is the first-level data cache, LLd is the last level of the data cache, and LL is the overall last-level cache.

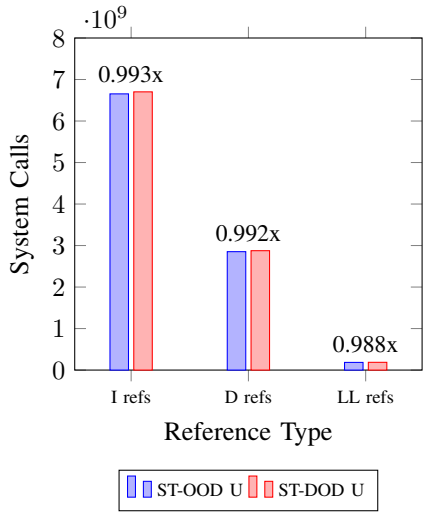
\begin{figure}[h!]
    \centering
    \begin{tikzpicture}
        \begin{axis}[
            ybar,
            bar width=7pt,
            symbolic x coords={I refs, D refs, LL refs},
            xtick=data,
            ymin=0, ymax=8000000000, 
            ylabel={System Calls},
            xlabel={Reference Type},
            width=0.7\linewidth, height=6cm, 
            scaled y ticks=true, 
            ytick={0, 1000000000, 2000000000, 3000000000, 4000000000, 5000000000, 6000000000, 7000000000, 8000000000},
            yticklabel style={/pgf/number format/precision=1},
            enlarge x limits={abs=0.75cm},
            legend style={at={(0.5,-0.3)}, anchor=north, legend columns=-1, font=\scriptsize},
            xticklabel style={font=\scriptsize, text width=1.2cm, align=center},
        ]
        \addplot coordinates {(I refs, 6653790377) (D refs, 2852624632) (LL refs, 184721576)};
        \addplot coordinates {(I refs, 6702522485) (D refs, 2876016747) (LL refs, 186839726)};
        \legend{ST-OOD U, ST-DOD U}
        \node at (axis cs:I refs, 6702522485) [above, font=\small] {0.993x};
        \node at (axis cs:D refs, 2876016747) [above, font=\small] {0.992x};
        \node at (axis cs:LL refs, 186839726) [above, font=\small] {0.988x};
        \end{axis}
    \end{tikzpicture}
    \caption{Single-thread: system calls.}
    \label{fig:sequential_cache_sc}
\end{figure}

The test was performed using 40,000 cells in the algorithm, and it can be observed that while the overall cache miss rates presented a high degree of similarity, DOD consistently demonstrated a slight advantage in critical cache levels, such as I1, indicating its inherent efficiency in data access patterns.

It is important to note that in Figure \ref{fig:sequential_cache_cu_rm}(\subref{fig:sequential_cache_cu}), the cache misses are measured in percentage. However, it is also essential to analyze the raw value of cache misses.

In Figure \ref{fig:sequential_cache_cu_rm}(\subref{fig:sequential_cache_rm}), the raw value of cache misses is displayed. It is possible to observe the great similarity in these raw values, with the most significant difference between them being 0.034 times in D1. Additionally, cache misses occur in large quantities, such as 114,226,091 recorded by OOD in I1, as seen in Figure 3 (b). However, even though they are similar, it is important to conclude that in the raw value of cache misses, the total sum in Figure 3 (b) resulted in 2,118,356 more cache misses for DOD. While this is a minimal difference for the machine, it still places OOD ahead of DOD in this aspect.

Conversely, Figure \ref{fig:sequential_cache_sc} presents the total number of instruction and data (read and write) calls for each of the two scenarios. There were 6,653,790,377 instruction calls in 'I refs' for OOD, 'D refs' represents data calls, and the last level cache calls are labeled as 'LL refs'. DOD ended up having more calls than OOD, albeit slightly, just as DOD had a higher raw value of cache misses than OOD in Figure \ref{fig:sequential_cache_cu_rm}(\subref{fig:sequential_cache_rm}). This makes sense as it maintains the proportionality shown in Figure \ref{fig:sequential_cache_cu_rm}(\subref{fig:sequential_cache_cu}) regarding the percentage of cache misses. Finally, in terms of calls, OOD had slightly fewer calls, giving it an advantage in this aspect.

\subsection{Multi-thread Execution}

In multi-thread execution, multiple threads work simultaneously to solve the algorithm.

\subsubsection{Execution Time}

\begin{figure*}[h]
    \centering
    \begin{minipage}[t]{0.48\textwidth}
        \centering
        \begin{tikzpicture}
            \begin{axis}[
                ybar,
                bar width=10pt,
                symbolic x coords={40000},
                xtick=data,
                ymin=0, ymax=40,
                ylabel={Execution Time [s]},
                xlabel={Number of Cells},
                width=6cm,
                height=7cm,
                ytick={0, 5, 10, 15, 20, 25, 30, 35, 40},
                enlarge x limits={abs=0.75cm},
                legend style={at={(0.5,-0.25)}, anchor=north, legend columns=-1},
            ]
            \addplot coordinates {(40000, 34.58)};
            \addplot coordinates {(40000, 30.32)};
            \legend{MT-OOD U, MT-DOD U}

            \node at (axis cs:40000, 34.58) [above, font=\small] {1.140x}; 
            \end{axis}
        \end{tikzpicture}
        \subcaption{Execution time graph obtained on Ubuntu system.}
        \label{fig:multi_thread_performance_et}
    \end{minipage}
    \hfill
    \begin{minipage}[t]{0.48\textwidth}
        \centering
        \begin{tikzpicture}
            \begin{axis}[
                ybar,
                bar width=10pt,
                symbolic x coords={40000},
                xtick=data,
                ymin=0, ymax=150,
                ylabel={Memory Used [MiB]},
                xlabel={Number of Cells},
                width=6cm,
                height=7cm,
                ytick={0, 25, 50, 75, 100, 125, 150},
                enlarge x limits={abs=0.75cm},
                legend style={at={(0.5,-0.25)}, anchor=north, legend columns=-1},
            ]
            \addplot coordinates {(40000, 84.36)};
            \addplot coordinates {(40000, 84.547)};
            \legend{MT-OOD U, MT-DOD U}

            \node at (axis cs:40000, 84.547) [above, font=\small] {0.998x}; 
            \end{axis}
        \end{tikzpicture}
        \subcaption{Memory usage graph obtained from the test conducted on Ubuntu system.}
        \label{fig:multi_thread_performance_mu}
    \end{minipage}
    \caption{Multi-thread: execution time and memory usage.}
    \label{fig:multi_thread_performance_et_mu}
\end{figure*}
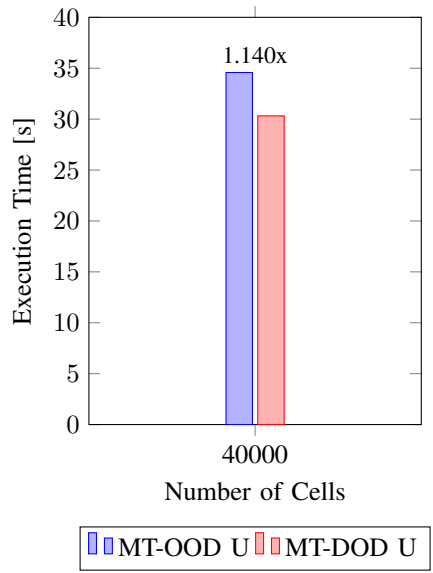
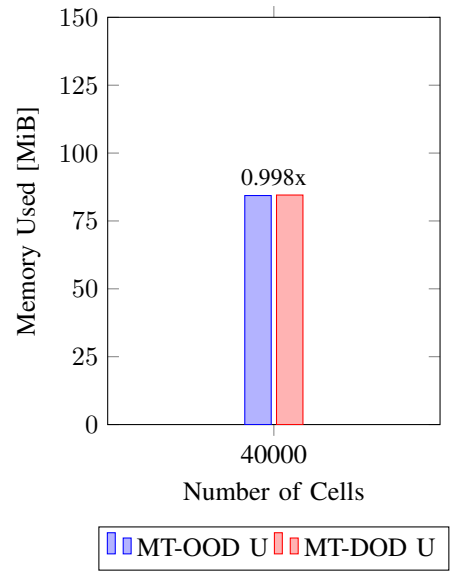

Figure \ref{fig:multi_thread_performance_et_mu}(\subref{fig:multi_thread_performance_et}) compares the multi-thread performance of the test program using DOD and OOD. The y-axis shows the execution time in seconds, while the values above the bars indicate the performance improvement of DOD compared to the OOD implementation.

As presented in the graph, DOD performs better than OOD in the multi-thread implementation. The execution time for the OOD code was 34.58 seconds, while for DOD, it was 30.32 seconds. The observed improvement was approximately 1.141 times when running the test program on Ubuntu 24.04.1 LTS with 4 threads using 40,000 cells.

\subsubsection{Memory Usage}

Figure \ref{fig:multi_thread_performance_et_mu}(\subref{fig:multi_thread_performance_mu}) compares the amount of memory used by the multi-thread test program using DOD and OOD. The y-axis shows the amount of memory used by the algorithm in mebibytes, while the values above the bars indicate the amount of memory used by the OOD implementation relative to DOD.

As shown in the graph, OOD performed slightly better than DOD in the multi-thread implementation. The amount of memory used by the OOD code was 84.36 MiB, while for DOD, it was 84.547 MiB. The observed reduction was approximately 0.998 times when running the test program on Ubuntu 24.04.1 LTS with 4 threads using 40,000 cells.

In the multi-thread experiments, an additional test was conducted to measure the number of threads used by the algorithm. This test indicated that both algorithms utilized between 1 and 3 threads out of the 4 available.

\subsubsection{Cache Utilization}

Figure \ref{fig:multi_thread_cache_cu_rm}(\subref{fig:multi_thread_cache_cu}) presents performance results related to cache misses in the multi-thread implementations. It highlights the difference in cache miss rates between DOD and OOD.

The y-axis shows the percentage of cache misses, while the values above the bars indicate the additional cache misses in the OOD implementation compared to DOD. The x-axis details the cache levels analyzed regarding cache misses, namely: I1, LLi, D1, LLd, and LL.

\begin{figure*}[h]
    \centering
    \begin{minipage}[t]{0.48\textwidth}
        \centering
        \begin{tikzpicture}
            \begin{axis}[
                ybar,
                bar width=7pt, 
                symbolic x coords={I1, LLi, D1, LLd, LL},
                xtick=data,
                ymin=0, ymax=4,
                ylabel={Cache Misses [\%]},
                xlabel={Cache Level},
                width=6cm, 
                height=6.5cm,
                ytick={0, 0.5, 1, 1.5, 2, 2.5, 3, 3.5, 4},
                enlarge x limits={abs=0.75cm},
                legend style={at={(0.5,-0.25)}, anchor=north, legend columns=-1, font=\small},
                xticklabel style={font=\small, text width=0.8cm, align=center},
            ]
            \addplot coordinates {(I1, 1.97) (LLi, 0) (D1, 3.1) (LLd, 0.1) (LL, 0)};
            
            \addplot coordinates {(I1, 2.02) (LLi, 0) (D1, 3.2) (LLd, 0.1) (LL, 0)};
            
            \legend{MT-OOD U, MT-DOD U}

            \node at (axis cs:I1, 2.02) [above] {\small 0.975x};
            \node at (axis cs:LLi, 0.0) [above] {\small 1x}; 
            \node at (axis cs:D1, 3.2) [above] {\small 0.969x};
            \node at (axis cs:LLd, 0.1) [above] {\small 1x};
            \node at (axis cs:LL, 0.0) [above] {\small 1x};
            \end{axis}
        \end{tikzpicture}
        \subcaption{Cache misses [\%] obtained on an Ubuntu system.}
        \label{fig:multi_thread_cache_cu}
    \end{minipage}
    \hfill
    \begin{minipage}[t]{0.48\textwidth}
        \centering
        \begin{tikzpicture}
            \begin{axis}[
                ybar,
                bar width=7pt, 
                symbolic x coords={I1, LLi, D1, LLd, LL},
                xtick=data,
                ymin=0, ymax=600000000,
                ylabel={cache misses},
                xlabel={Cache level},
                width=6cm, 
                height=6.5cm,
                ytick={0, 100000000, 200000000, 300000000, 400000000, 500000000, 600000000},
                enlarge x limits={abs=0.75cm},
                legend style={at={(0.5,-0.25)}, anchor=north, legend columns=-1, font=\small},
                xticklabel style={font=\small, text width=0.8cm, align=center},
            ]
            \addplot coordinates {(I1, 509788537) (LLi, 64601) (D1, 354630433) (LLd, 7120109) (LL, 7184710)};
            
            \addplot coordinates {(I1, 474137793) (LLi, 64104) (D1, 332908062) (LLd, 7143596) (LL, 7207700)};
            
            \legend{MT-OOD U, MT-DOD U}

            \node at (axis cs:I1, 509788537) [above] {\tiny 1.075x};
            \node at (axis cs:LLi, 64601) [above] {\tiny 1.008x};
            \node at (axis cs:D1, 354630433) [above] {\tiny 1.065x};
            \node at (axis cs:LLd, 7143596) [above] {\tiny 0.997x};
            \node at (axis cs:LL, 7207700) [above] {\tiny 0.997x};
            \end{axis}
        \end{tikzpicture}
        \subcaption{Bar chart of raw cache misses measured in the Ubuntu system.}
        \label{fig:multi_thread_cache_rm}
    \end{minipage}
    \caption{Multi-thread: cache misses and raw cache misses.}
    \label{fig:multi_thread_cache_cu_rm}
\end{figure*}
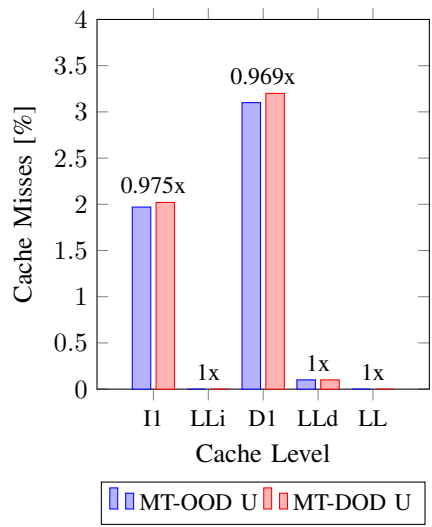
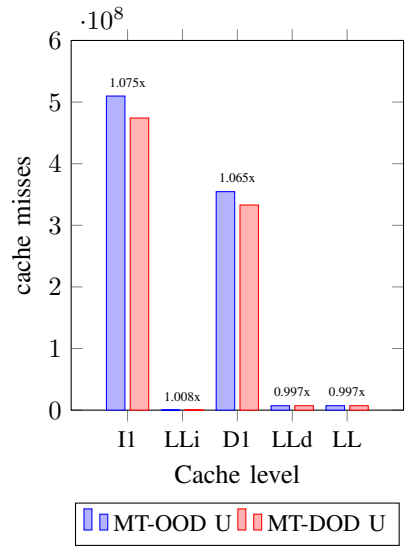

\begin{figure}[h!]
    \centering
    \begin{tikzpicture}
        \begin{axis}[
            ybar,
            bar width=7pt,
            symbolic x coords={I refs, D refs, LL refs},
            xtick=data,
            ymin=0, ymax=30000000000, 
            ylabel={System Calls},
            xlabel={Reference Type},
            width=0.7\linewidth, height=6cm, 
            scaled y ticks=true, 
            ytick={0, 5000000000, 10000000000, 15000000000, 20000000000, 25000000000, 30000000000},
            yticklabel style={/pgf/number format/precision=1},
            enlarge x limits={abs=0.75cm},
            legend style={at={(0.5,-0.3)}, anchor=north, legend columns=-1, font=\scriptsize},
            xticklabel style={font=\scriptsize, text width=1.2cm, align=center},
        ]
        \addplot coordinates {(I refs, 25845192653) (D refs, 11352625058) (LL refs, 864418970)};
        \addplot coordinates {(I refs, 23430033914) (D refs, 10257262700) (LL refs, 807045855)};
        \legend{MT-OOD U, MT-DOD U}
        \node at (axis cs:I refs, 25845192653) [above, font=\small] {1.103x};
        \node at (axis cs:D refs, 11352625058) [above, font=\small] {1.107x};
        \node at (axis cs:LL refs, 864418970) [above, font=\small] {1.071x};
        \end{axis}
    \end{tikzpicture}
    \caption{Multi-thread: system calls.}
    \label{fig:multi_thread_cache_sc}
\end{figure}
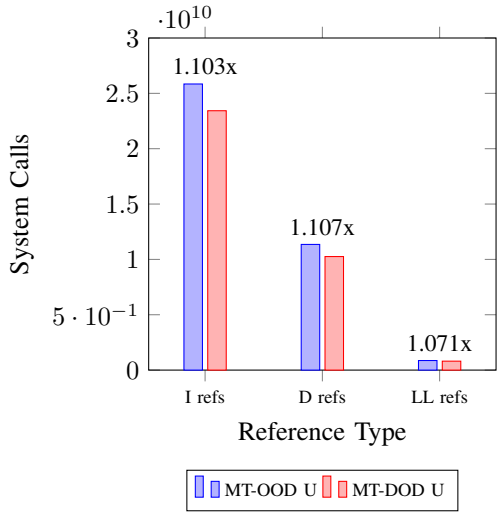

Once again, 40,000 cells were used in the algorithm, and it can be observed that DOD had approximately the same performance in terms of cache misses as OOD. In most cases, the cache miss percentage was the same for both, except in I1 and D1, where OOD performed better.

It is important to note that in Figure \ref{fig:multi_thread_cache_cu_rm}(\subref{fig:multi_thread_cache_cu}), the cache misses are measured as a percentage; however, it is also essential to analyze the raw values of cache misses.

Figure \ref{fig:multi_thread_cache_cu_rm}(\subref{fig:multi_thread_cache_rm}) presents the raw values of cache misses, while Figure \ref{fig:multi_thread_cache_sc} shows the total number of instruction and data (read and write) calls.

In Figure \ref{fig:multi_thread_cache_cu_rm}(\subref{fig:multi_thread_cache_rm}), the raw values of cache misses appear to be very similar, with the most significant difference being 0.075 times in I1. Additionally, cache misses occur in large quantities, such as 509,788,537 in OOD for I1, as shown in Figure \ref{fig:multi_thread_cache_cu_rm}(\subref{fig:multi_thread_cache_rm}). However, despite their similarities, it is essential to conclude that in terms of raw cache misses, DOD was superior in this regard, as it had a lower number of raw cache misses compared to OOD.

Figure \ref{fig:multi_thread_cache_sc} presents the number of calls for each of the two scenarios, such as 25,845,192,653 instruction calls in "I refs" for OOD. OOD had considerably more calls than DOD, which allowed DOD to outperform OOD in this aspect, as it was able to perform the same task with fewer system calls.

\subsection{Comparison between single-thread and multi-thread models}

As observed, the single-thread algorithm performed significantly better than the multi-threaded version. For instance, in OOD single-thread, the execution time was 2.981 seconds, whereas in OOD multi-thread, it took 34.58 seconds, resulting in a performance ratio of 11.6 times in favor of the single-thread version, which is a significant difference.

This discrepancy is also observed in execution times, the raw number of calls, and cache misses, with greater similarities between them in terms of memory usage and the percentage of cache misses. There are several reasons for this significant performance difference between both execution methods.

First, the overhead associated with thread management can be a critical factor. Although multi-threading is designed to better utilize multicore system resources, it also introduces additional costs. Each created thread needs to be managed, which includes operations such as creation, context switching, and synchronization. In tasks like the A* algorithm, which already involves a sequence of short and sequential operations, this overhead can outweigh the benefits of parallelism, resulting in lower performance.

Additionally, the A* algorithm utilizes data structures that are inherently dependent, such as priority queues. The need for synchronization when accessing these structures in a multi-threaded environment can create bottlenecks. For example, multiple threads attempting to insert or remove elements from the queue simultaneously require locks or mutual exclusion mechanisms, which reduce effective parallelism and increase execution time.

Another relevant factor is the problem’s granularity. A* processes small calculations per iteration, such as direction analysis. In such cases, the execution time of each operation is so short that the costs of dividing the work among threads and gathering the results outweigh the benefits of parallel processing. This is especially true when the problem does not scale sufficiently well to take advantage of all the processor's available cores.

In summary, because the implemented algorithm is relatively simple, with multiple short calculations, the cost of multi-thread management outweighed its potential advantages, and the single-thread method proved to be superior for these types of algorithms.

These findings, while highlighting the overheads inherent in parallelizing inherently sequential or fine-grained tasks like A* in simple maze scenarios, underscore a critical distinction for more complex applications. In such computationally intensive environments, where data volume and algorithmic complexity are substantially higher, the consistent, albeit sometimes marginal, performance benefits observed with Data-Oriented Design are expected to scale significantly. The principles of data contiguity and cache locality championed by DOD become paramount in these demanding contexts, suggesting that its architectural advantages would likely lead to more pronounced gains in system efficiency and overall throughput when applied to real-world, large-scale parallel problems.

\section{Conclusion}

Based on the obtained results, this study demonstrated that implementing Data-Oriented Design (DOD) principles can offer advantages over Object-Oriented Design (OOD) in terms of performance and cache utilization in multi-threaded CPUs, even in scenarios that do not involve high computational effort.

In sequential (single-threaded) tests, DOD showed slightly better performance than OOD in terms of execution time and memory usage. Optimizing data organization in DOD, aligning it with hardware memory architecture, minimized cache misses and maximized cache efficiency.

In multi-threaded tests, the superiority of DOD became more evident, with considerable gains in execution time and fewer raw calls and cache misses, whereas OOD showed slightly better performance in memory usage and cache miss percentage. DOD’s optimized data structure, emphasizing contiguity and efficient access, allowed for better utilization of multiple CPU cores, resulting in reduced processing time.

However, in simple algorithms with short calculations, the single-thread method outperformed the multi-threaded approach in both DOD and OOD. The overhead of thread management, the need for synchronization in dependent data structures, and problem granularity limited parallelism gains in low-processing scenarios. Therefore, for simple algorithms, single-threading tends to be more advantageous.

Regarding cache utilization, DOD consistently demonstrated a discernible advantage, particularly in minimizing cache misses. Although the percentage differences were modest in some tests, DOD's optimized data organization effectively enhanced cache spatial locality, leading to overall improved efficiency compared to OOD.

It is concluded that DOD consistently proved to be more efficient in multi-threaded CPU scenarios, even with low data volumes, showcasing its inherent strengths in optimizing hardware resource utilization. While the performance differences might appear subtle in these specific, simple algorithmic tests, they are indicative of a foundational architectural superiority that is expected to yield more substantial benefits in complex, large-scale applications. Therefore, while simpler algorithms might not fully expose the maximal benefits, the consistent advantages observed suggest that for performance-critical systems, particularly those involving complex AI algorithms or large datasets, Data-Oriented Design offers a superior approach to optimizing system performance and cache efficiency, guiding programmers towards more hardware-aware paradigms.

Re-evaluating the observed performance differences, it is crucial to recognize that even marginal improvements, as consistently demonstrated by DOD over OOD in our tests, hold significant implications for highly demanding applications and large-scale systems. While the A* algorithm, when implemented with simple data structures for maze solving, might not reveal drastic performance gaps, the consistent (albeit slight) advantages of DOD in execution time and cache efficiency highlight its inherent architectural benefits.

These small gains accumulate in complex scenarios with vast datasets and intensive computation, where every cycle and cache hit contributes to overall system responsiveness and resource utilization. The fundamental principles of data contiguity and spatial locality inherent in DOD are precisely what drive these subtle yet persistent improvements, offering a more hardware-friendly approach that scales effectively under increasing computational loads. Therefore, these findings, while not dramatically large in percentage for the chosen simple problem, serve as a strong indicator of DOD's potential for superior performance in more complex and data-intensive AI algorithms and parallel computing tasks.

Furthermore, future research could explore the application of Data-Oriented Design in real-world scenarios, particularly in fields where high-performance computing is crucial, such as gaming, scientific simulations, and financial modeling. Additionally, assessing the trade-offs between maintainability, readability, and performance when adopting DOD in large-scale software projects would provide a more comprehensive understanding of its practical implications. Lastly, empirical studies comparing developer productivity and debugging complexity between DOD and traditional programming paradigms could further clarify the feasibility of adopting DOD in diverse software development environments.

This study contributes to the advancement of knowledge on software optimization in multi-threaded systems, providing valuable insights for developers seeking to maximize the performance and efficiency of their applications.

\bibliographystyle{IEEEtran}
\bibliography{bibtex} 

@misc{nyberg2021investigating,
  title={Investigating the effect of implementing Data-Oriented Design principles on performance and cache utilization},
  author={Nyberg, Frank},
  year={2021}
}

@inproceedings{carvalho2002gap,
  title={The gap between processor and memory speeds},
  author={Carvalho, Carlos},
  booktitle={Proc. of IEEE International Conference on Control and Automation},
  volume={5000},
  number={10000},
  pages={15000},
  year={2002}
}

@article{nikolov2018oop,
  title={OOP is dead, long live Data-Oriented Design},
  author={Nikolov, Stoyan},
  journal={CppCon},
  year={2018}
}

@article{stroustrup1988object,
  title={What is object-oriented programming?},
  author={Stroustrup, Bjarne},
  journal={IEEE software},
  volume={5},
  number={3},
  pages={10--20},
  year={1988},
  publisher={IEEE}
}

@article{joshi2007data,
  title={Data-oriented architecture: A loosely-coupled real-time soa},
  author={Joshi, Rajive},
  journal={Whitepaper, Aug},
  year={2007}
}

@BOOK{culler1998,
  author    = {D. E. Culler and J. P. Singh and A. Gupta},
  title     = {Parallel Computer Architecture: A Hardware/Software Approach},
  publisher = {Morgan Kaufmann},
  year      = {1998},
}

@MISC{drepper2007,
  author    = {U. Drepper},
  title     = {What Every Programmer Should Know About Memory},
  year      = {2007},
  note      = {Retrieved from \url{https://people.freebsd.org/~lstewart/articles/cpumemory.pdf}},
}

@ARTICLE{lamport1978,
  author    = {L. Lamport},
  title     = {Time, Clocks, and the Ordering of Events in a Distributed System},
  journal   = {Communications of the ACM},
  volume    = {21},
  number    = {7},
  pages     = {558-565},
  month     = {July},
  year      = {1978},
}

@INPROCEEDINGS{amdahl1967,
  author    = {G. M. Amdahl},
  title     = {Validity of the Single Processor Approach to Achieving Large Scale Computing Capabilities},
  booktitle = {AFIPS Conference Proceedings},
  volume    = {30},
  pages     = {483-485},
  year      = {1967},
}

@INPROCEEDINGS{patterson1988,
  author    = {D. A. Patterson and G. Gibson and R. H. Katz},
  title     = {A Case for Redundant Arrays of Inexpensive Disks (RAID)},
  booktitle = {ACM SIGMOD International Conference on Management of Data},
  pages     = {109-116},
  year      = {1988},
}

@INPROCEEDINGS{dean2004,
  author    = {J. Dean and S. Ghemawat},
  title     = {MapReduce: Simplified Data Processing on Large Clusters},
  booktitle = {OSDI'04: Sixth Symposium on Operating System Design and Implementation},
  pages     = {137-150},
  year      = {2004},
}

@ARTICLE{larson2011,
  author    = {P. Larson and S. Blanas and C. Diaconu and C. Freedman and J. Patel and M. Zwilling},
  title     = {High-Performance Concurrency Control Mechanisms for Main-Memory Databases},
  journal   = {Proceedings of the VLDB Endowment},
  volume    = {5},
  number    = {4},
  pages     = {298-309},
  year      = {2011},
}

@INPROCEEDINGS{bienia2008,
  author    = {C. Bienia and S. Kumar and J. P. Singh and K. Li},
  title     = {The PARSEC Benchmark Suite: Characterization and Architectural Implications},
  booktitle = {Proceedings of the 17th International Conference on Parallel Architectures and Compilation Techniques},
  pages     = {72-81},
  year      = {2008},
}

@BOOK{martin2009,
  author    = {R. C. Martin},
  title     = {Clean Code: A Handbook of Agile Software Craftsmanship},
  publisher = {Prentice Hall},
  year      = {2009},
}

@BOOK{graham2004,
  author    = {P. Graham},
  title     = {Hackers \& Painters: Big Ideas from the Computer Age},
  publisher = {O'Reilly Media},
  year      = {2004},
}

@MISC{armstrong2003,
  author    = {J. Armstrong},
  title     = {Why OO Sucks},
  year      = {2003},
  note      = {Retrieved from \url{http://harmful.cat-v.org/software/OO\_programming/why\_oo\_sucks}},
}

@inproceedings{wingqvist2022,
  title={Evaluating the performance of object-oriented and data-oriented design with multi-threading in game development},
  author={Wingqvist, David and Wickstr{\"o}m, Filip and Memeti, Suejb},
  booktitle={2022 IEEE Games, Entertainment, Media Conference (GEM)},
  pages={1--6},
  year={2022},
  organization={IEEE}
}

@inproceedings{pellacini2019yocto,
  title={Yocto/GL: A data-oriented library for physically-based graphics},
  author={Pellacini, Fabio and Nazzaro, Giacomo and Carra, Edoardo and others},
  booktitle={ITALIAN CHAPTER CONFERENCE},
  year={2019},
  organization={The Eurographics Association}
}

@article{venancio2023desempenho,
  title={Desempenho do Design Orientado a Dados com Unity DOTS},
  author={Ven{\^a}ncio, Marcos Vin{\'\i}cius de Lima},
  year={2023}
}

@inproceedings{fontana2017exploiting,
  title={Exploiting cache locality to speedup register clustering},
  author={Fontana, Tiago Augusto and Almeida, Sheiny and Netto, Renan and Livramento, Vinicius and Guth, Chrystian and Pilla, La{\'e}rcio and G{\"u}ntzel, Jos{\'e} Lu{\'\i}s},
  booktitle={Proceedings of the 30th Symposium on Integrated Circuits and Systems Design: Chip on the Sands},
  pages={191--197},
  year={2017}
}

@book{NRC2011,
  author    = {National Research Council},
  title     = {The Future of Computing Performance: Game Over or Next Level?},
  year      = {2011},
  publisher = {The National Academies Press},
  address   = {Washington, DC},
  url       = {https://nap.nationalacademies.org/read/12980/chapter/5}
}

\end{document}